# Swarm Intelligence in Semi-supervised Classification

Shahira Shaaban Azab, Hesham Ahmed Hefny

*Abstract*— This Paper represents a literature review of Swarm intelligence algorithm in the area of semi-supervised classification. There are many research papers for applying swarm intelligence algorithms in the area of machine learning. Some algorithms of SI are applied in the area of ML either solely or hybrid with other ML algorithms. SI algorithms are also used for tuning parameters of ML algorithm, or as a backbone for ML algorithms. This paper introduces a brief literature review for applying swarm intelligence algorithms in the field of semi-supervised learning.

## I. INTRODUCTION

The amount of labeled data used for learning has a drastic effect on the performance of the classifier. However, labeled data is usually limited. Some approaches are proposed in the literature to enhance the performance of classifiers.

There are mainly three approaches[1]:
1. Exploit unlabeled data.
2. Use labeled data from different domains or problems.
3. Use feedback from of an oracle such as human expert.

Semi-supervised Learning (SSL)

SSL addresses the problem of rarely labeled data by using unlabeled data. SSL uses unlabeled data to identify the structure of the data. There are different models to solve the SSL problem such as:

1. Self-training: increase size of the training set by labeling unlabeled data. Then, the training set is enlarged by adding the most confident prediction to the training set.

2. Co-training: we have two classifiers. The features are represented into two disjoint subsets, and each of these subsets is sufficient to train a classifier. The most confident predictions of each classifier are added to the training set of the other classifier.

*Transfer learning* (TL)

In many machine learning approaches, there is an assumption that the data in the training sets and the data from the test sets are from the same domain and have the same distribution [2].

Shahira Shaaban Azab is with the Institute of statistical studies and research, Cairo University, Cairo; e-mail: Shahiraazazy@gmail.com.
Hesham Ahmed Hefney Shahira Shaaban Azab is with the Institute of statistical studies and research, Cairo University, Cairo.

However, knowledge from other domains can be of great help. For instance, the knowledge needed to recognize apples may help in recognition of pears. In general, humans can apply prior learned knowledge to solve new problems. They can transfer their knowledge to new domains. C++ programmers can learn similar programming languages such as Java fast[1]. Another example for the essential to transfer learning is the application with outdated data. Data collected may not follow the same distribution as the data collected in previous periods[3]. A popular application of transfer learning is the sentiment analysis. Reviews are classified into positive or negative reviews. With transfer learning, we can use reviews from similar products to train classifiers of other products[4].

Transfer learning does not use unlabeled data. However, it uses data from different domains[1]. The Chinese documents classification problem with small examples in the training set can use the knowledge of the English documents classification problem. Certainly, there must be some sort of bridge between the domains. For example, pairs of similar Chinese and -English documents.

In the area of the brain-computer interface, the performance depends on the amount of the available training data. However, the distribution of data varies according to the different subjects or even between the same subjects in different sessions. Hence, transfer learning can help[5].

*Active Learning*

Like SSL, Active learning uses both labeled and unlabeled to enhance classifier performance. It has many different names in literature such as query learning, selective sampling, sample selection, and experimental design in the statistics literature. In active learning, classifier selects the most relevant examples of the unlabeled data and use a human expert to label these data. Active learning assumes that the model can learn better with a few carefully chosen labeled data. Thus, it can ask an oracle such as human experts to label this small set of selected unlabeled data.

Figure 1 explains the effect of active learning on classification process. We have two classes A and B. The dashed line represents the decision boundary between the two classes (Figure 2 4 a). In (b) SVM classifier is used with the available seven examples. However, the limited number of training data leads to wrong decision boundary (the solid line). A more accurate decision boundary in (c) with SVM classifier using another seven examples chosen by an active learning process.

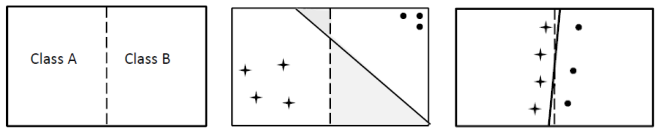

Figure 1 Active Learning[1].

Active learning asks the oracle to label some of the unlabeled data. There are different criteria to choose requested training patterns such as[1]:
- Uncertainty Sampling: Select the most uncertain labels.
- Query by Committee (QBC): Pick the training examples with the highest disagreement between committee of classifiers.
- Representativeness: select the most representative examples of the underlying data.

*1) Visual Learning*

Visual Learning makes use of users to help in classification process either by:
- Visual feedback of user helps in building a better training model. A clear example of visual learning is to use the help of users to choose best splitting points in a decision tree.
- Feedback from users to evaluate individual test instances.

## II. SWARM INTELLIGENCE (SI)

Swarm intelligence is defined as the field of computer science that designs efficient algorithms of collective intelligence. Self-organized, decentralized simple agents cooperate with no leader or supervisor to solve problems. SI inspired by the behavior of swarms, herds, flocks, or insect colonies that exhibit intelligent behavior to take consensus decisions. These individuals are working together towards achieving a goal. They depend on their collective knowledge. They exchange knowledge using chemicals (pheromone by ants), or dance (waggle dance by bees), or by broadcasting information (such as the global best in PSO and FA).

Ants are an obvious example of SI in nature. A single ant is weak, unintelligent and insignificant. For a single ant, there is no aim; a lonely worker ant roaming in circles until it dies of exhaustion[6][7]. A single ant has about 10,000 neurons in its brain. A very small number of neurons compared to humans who have about 100 billion neurons in their brain[7]. However, these helpless individuals are estimated to monopolize 15% of the mass of all land animals on Earth[8]. They adapt and survive more than what can be considered bigger and more intelligent creatures.

An example of SI in humans is developing of cooperative open source packages such as the development of the Linux operating system and Apache Software. In addition, the Wikipedia shows some sort of SI behavior.

SI is a popular research area that attracted the attention of researchers since its rise. SI algorithm is simple, and it can solve complex nonlinear problems and find an approximate solution. It is more appealing to find a near optimum solution in reasonable time. Meanwhile, classical methods are too expensive and difficult to implement.

Swarm intelligence, a term coined by Gerardo Beni. Then, the first two algorithms presented are ACO in 1992 and PSO in 1995. After the huge success of them, there was an exponential growth in the number of scientific works and research. Journals and conferences devoted to the swarm intelligence. SI now is a very active, fertile and promising research area. More and more algorithms are being designed[9].

### A. SI in nature

There are many examples of in nature [10][11][12][:
- Termites, which are a tiny insect, build a very complex nest structures sometimes as high as six meters with appropriate temperature and oxygen and carbon dioxide levels far beyond the ability of a single termite.
- Tasks are dynamically allocated within an ant colony, without any central manager.
- Leafcutter ants plant fungi and use it as food for their larvae.
- Bees can recruit other bees to help in gathering food by performing Tremble dance. Waggle dance in bee species allows them to share the information about the direction and distance to food sources and amount of food available there.
- Though, ants are very small (measuring only 2.2 to 2.6 mm in length), almost blind and have a very tiny brain. They use a great searching strategy and create highly sophisticated messaging systems. They can find the shortest path between the nest and food by secreting pheromone that other ant can smell it.
- Birds in a flock and fish in a school coordinate with their neighbors. Thus, they can fly/swim synchronously; they change speed and direction without any collision; they appear to move as a single unit.
- Predators, for example, a group of lionesses, have collaborative hunting strategies.
- Ants organize eggs in patterns where neighboring eggs have similar maturation times for more efficient feeding.

### B. Advantages of SI

There is an increase in the number of research papers about swarm intelligence. On average, there is an increase of more than 90% each year since 2000 according to study performed by[13] using Scopus database.

The main reasons for the popularity of SI is[14]:
- Using multiple interacting agents that evolve which mimics natural systems.
- Population-based approaches allow vector implementation which is a straightforward implementation.
- Simple, flexible and efficient.
- The ability to deal with complex problems in various fields.

### C. SI Challenges

*1) Understanding and Analyzing of algorithms*

There is no clear understanding of how algorithms learn, evolve, and solve problems. Why and how algorithms work.

The analysis of these algorithms is puzzling due to the heuristic nature of SI algorithms. The working mechanism of SI algorithms cannot be understood or explained clearly. A mathematical framework is needed to analyze the convergence and stability of SI algorithms[14].

"Still a basic question remains largely with no answer: Why does the ant algorithm work?" [15]. A lot of research for analyzing and understanding SI algorithms such as[16]. However, it is an open research issue till now.

Since understanding the working mechanism of SI algorithm has been unclear, choosing the appropriate algorithm for problems is random or by try and error. A mechanism is needed to select the most suitable algorithm for the problem at hand, taking into consideration available time and computational cost.

*2) Algorithm Parameters Tuning and Controlling*

Parameters setting of SI algorithms affects the performance of algorithms. Parameter tuning and controlling can be considered a hard optimization problem: finding the optimal parameters setting with optimal performance and a minimum computational cost. A popular approach for setting parameters is running algorithm using some trial values and choose the best setting to use in the application. Another approach is to use a well-established algorithm for optimizing parameters values of the new algorithm but how we can tune parameters of this well-established algorithm?[14], [17].

After parameters tuning and find optimal values for the parameters, Settings are fixed in all iterations of Algorithm with no reason. Parameters control means changing parameters value during iteration to achieve optimal performance of the algorithm[14].

*3) Global consensus*

There is no precise definition or terminology for SI algorithms. There are many variations and suggestions for every algorithm. There is no clear way to evaluate or elevate particular version of a certain algorithm.

*4) Randomness*

How much randomness needed for SI Algorithms? No randomness makes the algorithm deterministic and loses its ability to explore. The high amount of randomness turns algorithm to random search and lose its capability to exploit. Providing adequate randomness to enable the algorithm to explore without slowing convergence is crucial.

*5) High-Dimensionality*

Most real-world problems are complex, nonlinear, and large scale with thousands of design variables. Traditional problem-solving Methods like linear programming usually used to solve problems with half a million to several millions of design variables[14]. On the other hand, SI algorithms mostly used to solve moderate scale problems with hundreds of design variables. SI algorithms suffer from "Curse of Dimensionality". Studies show that the number of dimensions increased leads to the deterioration of the performance of SI algorithms. A metric for scalability of these algorithms is necessary. Methods to increase the efficiency of SI algorithms with complex, large-scale real-world problems are needed. There is no guarantee that algorithms work well for the problem of small or moderate size, can work with the same success in large scale real world problems.

## III. SWARM INTELLIGENCE IN SEMI-SUPERVISED CLASSIFICATION

In Literatures, there are few attempts to use SI Algorithms in SSL. Examples of SI algorithms in the domain of SSL are:

[18] reported a semi-supervised PSO with active learning for retrieval of images in large databases. In Content-Based Image Retrieval (CBIR), images are represented as features such as colors, textures, edges and the system retrieve the most similar images to the query images i.e. minimum distance to the required image from the image database. Since PSO is an efficient optimization algorithm, the authors depict retrieval problem as an optimization problem. Then, PSO is used to minimize weighted Euclidean distance between the query image and images in the databases. Images are represented as a features vector. PSO particles were encoded as query points in the multidimensional solution space. Features of images were represented as a discrete set of points. The global best particle was encoded as the query image. The particles are moving in the search space looking for the most similar feature to the query image. The best solutions are shown to the user. User feedback at each stage guides the search process. The algorithm is terminated when resulted images satisfy user or when the required number of images retrieved, or when a predefined number of iterations is reached. The results of the algorithm are evaluated according to the user feedback in each iteration. The authors used only a subset of features, namely colors and text. Experiments conducted using a Caltechimage-256 image database. The results showed the effectiveness of the reported PSO.

[19], [20] presented a self-training semi-supervised algorithm based on the aggregation pheromone behavior of ants. In APSSC, Labeled and unlabeled data points are represented as ants. Labeled ants can secrete pheromones. Each class represents an ant colony, and each labeled ant secrets pheromone of type c where c is the class label of the ant. The unlabeled ants smell accumulated pheromone and choose the class with highest concentration pheromone. The class of an ant is updated over time according to the accumulated value of the pheromone. In the first stage of the training, the classifier trained using labeled data only. After that, the trained classifier used to predict labels for the unlabeled data. Then, the most confident unlabeled data with their predicted labels are added to the training set which is represented as colonies of ants. The newly created training set is used to train the classifier. This process is repeated until ant colony is well-formed (centers of colonies do not change in two successive iterations). In the testing phase, the test data points (ants) are assigned to the colony with the highest average of the aggregation pheromone density. The proposed classifier is compared with two supervised classifiers (multilayer perceptron and support vector machine) and three semi-supervised classifiers (semi-supervised classification by low-density separation and concave-convex procedure for transductive support vector

machine, self-training semi-supervised support vector machine). In the experiments, four non-linear separable two dimension artificial data sets and five benchmark data sets are used. The results assure that the performance of the algorithm is promising.

[21] proposed semi-supervised self-training PSO classifier for Chinese text categorization. The goal of the proposed classifier is predicting the topics or subjects of documents. The author assumed that each document belongs to exactly one class. The experiments were conducted on a Chinese text corpus which has ten categories and 950 news articles. The proposed PSO classifier used each document's previous prediction and neighbors' information to predict category of documents. In the first stage, the k weighted nearest neighbors classifier was used to classify unlabeled data. For each test data point x, the distance between x and all training examples was calculated and the set k of the closest training data points were selected. The prediction for the class of point x is chosen according to the majoring voting of k neighbors. x could be added to an archive with random probability. Then, for a fixed number of iteration, the distance between x and archive data was calculated and the predicted label for x was modified. If a closer neighbors found, the archive data are updated. The proposed classifier was compared with the k nearest neighbors, the k weighted nearest neighbors, and the self-learning classifier using weighted nearest neighbors. According to the authors, the performance of the presented PSO- k nearest neighbors semi-supervised classifier is promising.

[22] used graph-based transductive learning methods using random walk ACO. In the proposed ACO, labeled data are represented as ants in different nests, each class is an ant colony, and unlabeled data are food sources. Each colony secretes unique types of pheromone. All data points were represented as nodes in the graph. The colonies competed for unlabeled data (food sources). The proposed approach used the idea of label propagation where labeled nodes propagate their labels to unlabeled nodes in the graph. The random walk of the ants from the same nest for exploring food sources around them is a tree. The pheromone evaporated during the moves of the ants around the graph. Thus, the goal of the proposed algorithm is to find the maximum spanning tree and with maximum transition probability and the pheromone intensity. The experiments used ten real-world data sets. The results were compared in terms of classification accuracy with three algorithm Nearest Neighbor, Multi-level Component Propagation, μcAntMiner. The results show the algorithm is efficient.

[23] proposed a dynamic self-training semi-supervised classifier based on swarm intelligence for Evolutionary data. Evolutionary data changes with time such as weblogs and GPS sensors. Unlike streaming data, there is no continuous flow of data. Each class was represented as an ant colony secreting different types of pheromone. The unlabeled ants choose their colony based on secreted pheromone. The unlabeled ant moves to its nearest nest using k-Nearest Neighbor. The algorithm tested using three data sets: the artificial two-moons, Mushroom, and Hyperplane datasets from the UCI repository. The authors reported the effectiveness of the proposed algorithm.

[24] proposed "Ant-Labeler" a semi-supervised self-training algorithm. Ant-Labeler used ACO as a self-learning wrapper for labeling unlabeled data. The authors reported that Ant-Labeler ACO outperforms state of the art algorithm in half of the used datasets and the performance of the proposed algorithm improves as the number of labeled instance decreases. The results of Ant-Labeler compared with four algorithms: cS3VM, self-learning C4.5, APSSC in [Halder, A. et al., 2010], and the original version of cAnt-MinerPB. 1%, 10%, 40% and 70% of label data on 20 data sets from UCI repository.

[25] presented a semi-supervised model using cluster and label approach. The authors used a local best version of Particle swarm optimization PSO to cluster data. The proposed model create a spherical cluster. However, the cluster may not be spherical. The guidance of labels helps to reshape the formed clusters according to the available supervised information. Thus, SPSO with the help of available labeling information achieves competitive results compared to traditional semi-supervised algorithms. In addition, SPSO makes use of the nearest neighbor to resolve the conflicts of mixed clusters.